\documentclass{tlp}
 
\usepackage{latexsym}
\usepackage{url}
\usepackage{aopmath}
 
\title[Theory and Practice of Logic Programming]
{On the Expressibility of Stable Logic Programming}

\author[V.W. Marek and J.B. Remmel]
{Victor W. Marek \\
Department of Computer Science, University of Kentucky, Lexington, KY 40506,
USA \\
\email{marek@cs.uky.edu}
\and
Jeffrey B. Remmel\\
Department of Mathematics, University of California, San Diego, La Jolla, 
CA 92093, USA\\
\email{jremmel@ucsd.edu}}

\pagerange{\pageref{firstpage}--\pageref{lastpage}}

\newcommand{\la}{\langle}
\newcommand{\ra}{\rangle}
\begin{document}

\label{firstpage}
 
\volume{\textbf{10} (3):}
\setcounter{page}{1}
\maketitle

\begin{abstract}
Schlipf \cite{sch91} proved that Stable Logic Programming (SLP) solves all
$\mathit{NP}$ decision problems. We extend Schlipf's result to prove
that SLP solves all search problems in the class $\mathit{NP}$.
Moreover, we do this in a uniform way as defined in \cite{mt99}.
Specifically, we show that there is a single $\mathrm{DATALOG}^{\neg}$
program $P_{\mathit{Trg}}$ such that given any Turing machine $M$,
any polynomial $p$ with non-negative integer coefficients and any
input $\sigma$ of size $n$ over a fixed alphabet
$\Sigma$, there is an extensional database $\mathit{edb}_{M,p,\sigma}$
such that there is a one-to-one correspondence between the stable models of
$\mathit{edb}_{M,p,\sigma} \cup P_{\mathit{Trg}}$ and
the accepting computations of
the machine $M$ that reach the final state in at most $p(n)$ steps.
Moreover, $\mathit{edb}_{M,p,\sigma}$ can be computed in polynomial time from
$p$, $\sigma$ and the description of $M$ and the decoding of such accepting
computations from its corresponding stable model of
$\mathit{edb}_{M,p,\sigma} \cup P_{\mathit{Trg}}$
can be computed in linear time. A similar statement holds for
Default Logic with respect to $\Sigma_2^\mathrm{P}$-search
problems\footnote{The proof of this result involves additional
technical complications and will be a subject of another publication.}.
\end{abstract}

\section{Introduction} \label{intro}
The main motivation for this paper comes from recent developments
in Knowledge Representation, especially the appearance of a
new generation of systems \cite{cmt96,ns96,elmps97} based
on the so-called Answer Set Programming (ASP) paradigm
\cite{nie98,cp98a,mt99,li98}. The emergence of these ASP systems suggest
that we need to revisit one of the basic issues in the foundations of
ASP, namely, how can we characterize what such ASP systems can
theoretically compute. 

Throughout this paper, we shall focus mostly on one
particular ASP formalism, specifically, the Stable Semantics for Logic
Programs (SLP) \cite{gl88}. We note that the underlying methods of ASP
are similar to those used in Logic Programming \cite{ap90} and
Constraint Programming \cite{jm94,ms99}. That is, like Logic
Programming, ASP is a declarative formalism and the semantics of all
ASP systems are based on logic. Like Constraint Programming, certain
clauses of an ASP program act as {\em constraints}. There is a
fundamental difference between ASP programs and Constraint Logic
programs, however. That is, in Constraint Programming, the constraints
act on individual elements of Herbrand base of the program while
the constraint clauses in ASP programs act more globally in that they place
restrictions on what subsets of the Herbrand base can be acceptable
answers for the program. For example, suppose that we have a problem
$\Pi$ whose solutions are {\em subsets} of some Herbrand base $H$. In
order to solve the problem, an ASP programmer essentially writes a
logic program $P$ that describes the constraints on the subsets of
$H$ which can be answers to $\Pi$. The basic idea is that the
program $P$ should have the property that there is an easy decoding
of solutions of $\Pi$ from stable models of $P$ and that all solutions
of $\Pi$ can be obtained from stable models of $P$ through this
decoding. The program $P$ is then submitted to the ASP engine such as
{\em smodels} \cite{ns96}, {\tt dlv} \cite{elmps97} or DeReS
\cite{cmt96} which computes the stable models of the program $P$.
Thus the ASP engine finds the stable models of the program (if any
exists) and one reads off the solutions to $\Pi$ from these stable
models. Notice that the idea here is that all solutions are equally
good in the sense that any solution found in the process
described above is acceptable. Currently, the systems based on ASP
paradigm are being tested on the problems related to planning
\cite{lif99,nie98}, product configuration \cite{sn99}, combinatorial
optimization problems \cite{cmmt99,nie98} and other domains. 

It is a well-known fact that the semantics of existing Logic Programming
systems such as Prolog, Mercury and LDL have serious problems,
principally due to necessary compromises in the implementations. For
instance, the unification algorithms used by most dialects of Prolog do
not enforce the occurs check and hence these systems can produce
incorrect results \cite{ap94}. Moreover, the processing strategies
of Prolog and similar languages have the effect that correct logic
programs can be non-terminating \cite{ap93}. While good programming
techniques can overcome these problems, it is clear that such deficiencies
have restricted the appeal of the Logic Programming systems for ordinary
programmers and system analysts. The promise of ASP and, in particular,
of SLP and its extensions, such as Disjunctive Logic Programming
\cite{gl91,elmps97}, is that a new generation of logic programming systems
can be built which have a clear semantics and are easier to program
than the previous generation of Logic Programming systems. In
particular, both of the problems referred to above, namely, the
occurs check problem and the termination problem, do not exist in SLP.
Of course, there is a price to pay, namely, SLP systems only accept
programs without function symbols. Consequently, one of the basic data
structures used in Prolog, the {\em term}, is not available
in SLP. Thus SLP systems require the programmer to explicitly
construct many data structures.
In SLP programming, predicates are
used to construct the required data structures and clauses that serve as
constraints are used to ensure that the predicates behave properly with
respect to semantics of the program. SLP programs are always
terminating because the Herbrand base is finite and hence there are
only a finite number of stable models. In addition, unlike the case of
usual Logic Programming, the order of the clauses
of the program does not affect the set of stable models of the 
program\footnote{However it is the case that the order of clauses can
affect the processing time of the ASP engine.}. Finally the stable
semantics of logic programs is well understood so that SLP programs have
clear semantics.

The restriction that ASP programs do not contain 
function symbols is crucial. First, it is well known that
once one allows function symbols in a logic program $P$, the
Herbrand base becomes infinite.
Moreover, the stable models of logic programs {\em with}
function symbols can be immensely complex. For example,
for stratified logic programs \cite{abw87,przy86}, the perfect model is the
unique stable model of that program \cite{gl88}. Apt and Blair \cite{ab89}
showed that perfect models of stratified logic programs
capture precisely the arithmetic sets. That is, they show that for
a given arithmetic set $X$ of natural numbers, there is a finite
stratified logic program $P_X$ such that in the perfect model of $P_X$, some
predicate $p_X$ is satisfied by precisely the numbers in $X$. This was the 
first result that showed that
it is not possible to have meaningful practical programming
with general stratified programs {\em if we allow} unlimited use
of function symbols.
The result of
\cite{ab89} was extended in \cite{bms91} where Blair, Marek, and Schlipf
showed that the set of stable models of a locally stratified program can
capture any set in the hyperarithmetic hierarchy. Marek, Nerode,
and Remmel \cite{mnr94} showed that the problem of finding a stable model 
of a finite (predicate) logic program $P$ is essentially equivalent to 
finding a path through an infinite branching recursive tree. 
That is, given an
infinite branching recursive tree $T \subseteq \omega^{< \omega}$, there is 
a finite program $P_T$ such that there is a one-to-one
degree-preserving correspondence between the infinite paths through
$T$ and the stable models of $P_T$ and, vice versa, given a finite
program $P$, there is a recursive tree $T_P$ such that there is
one-to-one degree preserving correspondence between the stable models
of $P$ and the infinite paths through $T_P$. One consequence of this
result is that the problem of determining whether a finite predicate
logic program has a stable model is $\Sigma_1^1$-complete.
More results on the structure of the family of stable models of 
programs can be found in \cite{cr99}. 

All the results mentioned in the previous paragraph show that the stable model 
semantics for logic programs admitting function symbols can be used 
practically,  only
in a very limited setting. XSB system attempts to do deal with this problem 
by computing only the well-founded semantics. When the 
well-founded semantics is 
total, the resulting model is the unique stable model of the program. 
Unfortunately, the class of programs for which it succeeds is not
intuitive \cite{ssw97}. Yet another attempt to return 
the power of function symbols to the language
has been made in \cite{bo01}. The class of programs considered in 
\cite{bo01} allows one to express recursively
enumerable sets, but not more complex sets so that, at best, one could 
get a more compact representation of problems solved with ordinary Prolog.

As stated above, ASP systems propose a
more radical solution to the problem of complexity of stable models of logic
programs with function symbols, namely, abandoning function symbols
entirely. Once this is accepted, the semantics of a logic program $P$ can be
defined in two stages. First, we assume, as in standard Logic Programming,
that we interpret $P$ over the Herbrand universe of $P$ determined by the
predicates and constants that occur in $P$. Since the set of constants
occurring in the  program is finite, we can ground the program
in these constants to obtain a finite
propositional logic program $P_{\mathit{g}}$. The stable models of $P$ are
by definition the stable models of $P_{\mathit{g}}$. The process
of grounding is performed by a separate grounding engine such as
{\em lparse} \cite{ns96}. The grounded program $P_{\mathit{g}}$ is
then passed to an engine that computes 
stable models of propositional logic programs. It is then easy to check that the features of SLP mentioned
above, i.e., the absence of occurs check and termination problems and
the independence of the semantics from the ordering of the clauses of
the program, automatically hold.
That is, since grounding uses only very
limited part of unification, the occurs check problem is eliminated.
The space of candidates for stable models is finite and so there is
no termination problem. Finally, the stable semantics of propositional
programs does not depend on the order of clauses.

The language of logic programming without function symbols was
studied by the database community with the hope that it could lead to
new, more powerful, database language \cite{ul88}. This language is
called $\mathrm{DATALOG}^{\neg}$
and some database systems such as DB2 implement the positive part of 
$\mathrm{DATALOG}^{\neg}$.
The fact that admitting negation in the bodies of clauses leads to
multiple stable models was unacceptable from the database perspective.
Hence the database community preferred other semantics for 
$\mathrm{DATALOG}^{\neg}$  programs 
such as the well-founded semantics \cite{vrs91}
or the inflationary semantics \cite{ahv95}.

The main purpose of this paper is to revisit the question of what can
be computed by logic programs without functions symbols under the stable
model semantics. First, consider the case of finite propositional
programs. Here the situation is simple. Given a set $At$ of propositional
atoms, let $\cal F$ be a finite antichain of subsets of $At$, i.e. whenever
$X,Y \in {\cal F}$, $X \subseteq Y$, then $X = Y$. Then one can show that
there is a logic
program $P_{\cal F}$ such that $\cal F$ is precisely the class of all stable
models of $P_{\cal F}$ \cite{mt93}. Moreover, the family of stable models
of any program $P$ forms such an antichain. Thus in the case of
finite propositional logic programs, we have a complete
characterization of the possible sets of stable models. However, this
result by itself does not tell us anything about the 
uniformity and the effectiveness
of the construction. The basic complexity result for SLP propositional
programs is due to Marek and Truszczy\'{n}ski \cite{mt88} who
showed that the problem of deciding whether a finite propositional logic
program has a stable model is $\mathit{NP}$-complete.
For $\mathrm{DATALOG}^{\neg}$,  an analogous result has been 
obtained in \cite{sch91}.

To precisely formulate our question about what can be computed by
logic programs without functions symbols under the stable model semantics,
we first need to recall the notion of {\em search problem} \cite{gj79} and
of a {\em uniform} logic program \cite{mt99}. A search problem is a set
$\cal S$ of finite instances such that, given
any instance $I \in {\cal S}$, there is a set $S_I$ of solutions to
$\cal S$ for the instance $I$. It is possible that for some 
instances $I$, $S_I$ is the empty set. 
For example, the search problem
may be to find Hamiltonian paths in a graph. Thus, the set of
instances of the problem is the set of all finite graphs. Then,
given any instance, i.e. a graph $G$, $S_G$ is the set of all 
Hamiltonian paths of $G$. We say that an 
algorithm solves a search problem $\cal S$ if it returns a
solution $s \in S_I$ whenever $S_I$ is non-empty and it returns the string
``empty'' otherwise.
We say that a search problem $\cal S$ {\em is in $\mathit{NP}$} if there is
such an algorithm which
can be computed by a non-deterministic polynomial time Turing machine.
We say that search problem $\cal S$ {\em is solved by a uniform logic program}
if there exists a single logic program $P_{\cal S}$, a polynomial time 
extensional data base transformation function  $edb_{\cal S}$  
and a polynomial time solution decoding function 
$sol_{\cal S}(\cdot,\cdot )$ such that for every instance $I$ in $\cal S$,
\begin{enumerate}
\item $edb_{\cal S}(I)$ is a finite  set of facts,
i.e. clauses with empty bodies and no variables, 
\item  whenever $sol_{\cal S}(I)$ is non-empty,  $sol_{\cal S} (I,\cdot)$ maps
the set of stable models of the $edb_{\cal S}(I) \cup P_{\cal S}$
onto the set of solutions $S_I$ of $I$ and 
\item if $sol_{\cal S}(I)$ is empty,   then 
$edb_{\cal S}(I) \cup P_{\cal S}$ has no stable models.
\end{enumerate}

We note that decision problems can be viewed as special cases of
search problems. Schlipf \cite{sch91} has shown that the class
of {\em decision} problems in $\mathit{NP}$ is captured precisely by uniform
logic programs. Specifically he proved that a decision problem is
solved by a uniform logic program if and only if it is in $\mathit{NP}$.
An excellent review of the complexity and expressibility results for Logic
Programming can be found in \cite{degv01}.
The goal of this paper is to prove a strengthening of Schlipf's result as
well as prove a number of related facts. 
We will prove that Schlipf's result can be extended to all $\mathit{NP}$
{\em search}
problems. That is, we shall show that there is a
single logic program $P_{\mathit{Trg}}$ that
is capable of simulating polynomial time nondeterministic Turing
machines in the sense that given any polynomial time nondeterministic Turing
machine $M$, any input $\sigma$, and any run-time polynomial $p(x)$,
there is a set of facts $edb_{M,p,\sigma}$ 
such that a stable model of $P_{\mathit{Trg}} \cup edb_{M,p,\sigma}$ codes
an accepting computation of $M$ started with input $\sigma$ that terminates
in $p(|\sigma|)$ or fewer steps
and any such accepting computation of $M$ is coded
by some stable model of $P_{\mathit{Trg}} \cup edb_{M,p,\sigma}$.
This result will show
that logic programs without function symbols under the stable model
semantics capture all $\mathit{NP}$-search
problems\footnote{As pointed by M. Truszczy\'nski, for our goal of describing 
the complexity of the Stable Logic Programming, a weaker
result is sufficient. That is, we need only show that
for each instance $I$ of an $\mathit{NP}$ search problem $\Pi$, there is a
program $P_I$ and a polynomial time projection from the collection of
stable models of $P_I$ to the set of solutions of $I$. Our result shows
that this property holds in a stronger form. Namely, there is
a single program with a varying
extensional database.}. The converse implication, that is, a search problem
computed by a uniform logic program $P$ is an $\mathit{NP}$-search problem
is obvious since one can compute a stable model $M$ of a program
by first guessing $M$ and then doing a polynomial time check to verify that 
$M$ is a stable model of the program.

\section{Technical preliminaries}\label{techprel}
In this section we shall formally introduce several notions that will be
needed for the proof of our main result. The proof of
our main  result uses the basic idea used by
Cook \cite{co71} in his proof of the $\mathit{NP}$-completeness of the 
satisfiability problem.
First, we introduce the set of logic programs that we will study.
We will consider here only so called $\mathrm{DATALOG}^{\neg}$ programs.
Specifically, a clause is an expression of the form
\begin{equation} \label{eq.1}
p(\overline{X}) \leftarrow q_1(\overline{X}), \ldots, q_m(\overline{X}),
\neg \ r_1(\overline{X}), \ldots, \neg\ r_n(\overline{X})
\end{equation}
where $p,q_1,\ldots,q_m,r_1,\ldots,r_n$ are atoms, possibly with variables
and/or constants. Here we abuse notation by writing 
$p(\overline{X})$ to mean that the variables that occur in 
the predicate $p$ are contained in $\overline{X}$. A program 
is a finite set $P$ of clauses of the form
(\ref{eq.1}). We assume that the underlying language ${\cal L}_P$ of any given program $P$ is determined by the constants and predicate symbols which occur in the program. Thus the  Herbrand universe $U_P$ of $P$ is just the 
set of all constant terms occurring in $P$ and the Herbrand base
$H_P$ of $P$ is the set of all ground atoms of the language ${\cal L}_P$.
Since there are no function symbols in our programs, both the Herbrand
universe and the Herbrand base of the program are finite.

A ground instance of a clause $C$ of the form (\ref{eq.1}) is the result
of a simultaneous substitution of constants $\overline{c}$ for variables 
$\overline{X}$ occurring in
$C$. Given a program $P$, $P_g$ is the propositional program consisting
of all ground substitutions of the clauses of $P$.
Given a propositional program $P$ and a set $M$ included in its Herbrand
base, $H_P$, the
Gelfond-Lifschitz transformation of $P$ relative to  $M$ 
is the program $GL(P,M)$ arising from $P$ as follows. First, eliminate
all clauses $C$ in $P$ such that for some $j$, $1\leq j\leq n$, 
$r_j(\overline{c}) \in
M$. Next, in any remaining clause, eliminate all negated
atoms. The resulting set of clauses forms a program, $GL(P,M)$, which is
a Horn program and hence it possesses a least model $N_M$. We
say that $M$ is a {\em stable model of the propositional program}
$P$ if $M = N_M$. Finally, given any program $P$ with variables, we say that $M$ is a stable model of a
program $P$ if $M$ is a stable model of
the propositional program $P_g$.

A nondeterministic Turing Machine is a septuple of the form
\[
M = (Q, \Sigma, \Gamma, D, \delta, s_0, f).
\]
Here $Q$ is a finite set of states and $\Sigma$ is a finite alphabet
of input symbols. We assume $Q$ always contains two special states, $s_0$, the
start state, and $f$, the final state.
We also assume that there is a special symbol $B$ for ``blank'' such
that $B \notin \Sigma$. The set
$\Gamma = \Sigma \cup \{B\}$ is the set of tape symbols.
The set $D$ is the set of move directions consisting  of the elements 
$l,r,$ and
$\lambda$ where $l$ is the ``move left'' symbol, $r$ is the ``move right'' 
symbol and $\lambda$ is the ``stay put'' symbol. The function $\delta :
Q\times \Gamma \rightarrow {\cal P} (Q \times \Gamma \times D)$ is the
transition function of the machine $M$. Here 
${\cal P} (Q \times \Gamma \times D)$ denotes the power set of 
the set $Q \times \Gamma \times D$.  We can also think of $\delta$ as
a 5-ary relation.  Thus we can represent the transition function
of the machine $M$ as a collection of atoms describing 5-tuples.
We assume $M$ operates on a one-way infinite tape where
the cells of the tape are labeled from left to right by $0,1, 2, \ldots$. 
To visualize the behavior of
the machine $M$, we shall talk about the read-write head of the
machine. At any given time in a computation, the
read-write head of $M$ is always in some state $s \in Q$ and
is reading some symbol $p \in \Gamma$ which is in a cell $c$ of the tape. 
It then picks an instruction
$(s1,p1,d) \in \delta(s,p)$ and then replaces the symbol $p$ by $p1$, changes
its state to state $s1$, and moves according to $d$.  We assume that 
at the start of the computation of $M$ on input $\sigma = (\sigma(0), 
\ldots, \sigma(n-1))$, cells $0, \ldots, n-1$ contain the symbols 
$\sigma(0), \ldots, \sigma(n-1)$ respectively and all cells to the 
right of cell $n-1$ are blank. We do not impose (as it is often done)
any special restrictions on the state of the tape and the position of
the head at the end of computation. However, we assume that at the 
start of any computation, the read-write head is in state $s_0$ and is 
reading the symbol in cell 0.

Suppose we are given a Turing machine $M$ whose runtimes are bounded by a
polynomial $p(x) = a_0 + a_1x+ \cdots + a_k x^k$
where each $ a_i \in N = \{0,1, 2, \ldots \}$ and $a_k \neq 0$. That is,
on any input of size $n$, an accepting computation terminates in at most
$p(n)$ steps. Then any accepting computation on input $\sigma$
can affect at most the first $p(n)$ cells of the
tape. Thus in such a situation, there is no loss in only considering
tapes of length $p(n)$. Hence in what follows,
one shall implicitly assume that the tape is finite.
Moreover, it will be convenient to modify
the standard operation of $M$ in the following ways.\\
1. We shall assume $\delta(f,a) = \{(f,a,\lambda)\}$ for all $a \in
\Gamma$.\\
2. Given an input $x$ of length $n$, instead of immediately halting
when we first get to the final state $f$ reading a symbol $a$,
we just keep executing
the instruction $(f,a,\lambda)$ until we have completed $p(n)$ steps.
That is, we remain in state $f$, we never move, and we never
change any symbols on the tape after we get to state $f$.\\
The main effect of these modifications is that all accepting computations will
run for exactly $p(n)$ steps on an input of size $n$. 

\section{Uniform coding of Turing Machines by a Logic Program}\label{tmcode}

In this section, we shall describe the logic program $P_{\mathit{Trg}}$ and
our extensional data base function $edb_{M,p,\sigma}$ described above.
The key to our construction is the fact that at any given moment of time, the
behavior of a Turing machine $M$ depends only on the current state of tape, 
the
position of the read-write head and the set of available instructions.
Our coding of Turing machine computation reflects this simple observation.
First, we define the language (i.e. a signature) of the program
$P_{\mathit{Trg}}$. The set of predicates that will occur in our extensional
database are the following:\\
$time(X)$ for ``$X$ is a time step'',\\
$cell(X)$ for ``$X$ is a cell number'',\\
$symb(X)$ for ``$X$ is a symbol'',\\
$state(S)$ for ``$S$ is a state'',\\
$i\_position(P)$ for ``$P$ is the initial position of the read-write
head'',\\
$data(P,Q) $ for ``Initially, the tape stores the symbol $Q$ at
the cell $P$'',\\
$delta(X,Y,X1,Y1,Z)$ for ``the triple $(X1,Y1,Z)$ is an executable
instruction when the read-write head is in state $X$ and is reading the 
symbol  
$Y$'' (thus $delta$ represents the
five-place relation $\delta$),\\
$neq(X,Y)$ for ``$X$  is different from $Y$'' \\
$eq(X,Y)$ for ``$X$ is equal to  $Y$''\footnote{Technically, we should use 
a separate  equality and inequality relation for each type, but 
we will not do so.},\\
$succ(X,Y)$ for ``$Y$ is equal to $X+1$''\footnote{For the clarity 
of presentation we will use equality symbol $=$, inequality symbol, 
$\neq$ and relation described by the successor
function $+1$, instead of $\mathit{eq}, \mathit{neq}$, and 
$\mathit{succ}$.}

Fix a nondeterministic Turing machine $M = (Q,\Sigma,\Gamma, D, 
\delta, s_0,f)$, a run-time polynomial $p(x)$ and an input 
$\sigma = (\sigma(0), \ldots, \sigma(n-1))$ of length $n$.  
This given, we now
define the extensional database $ext_{M,p,\sigma}$. First,
$ext_{M,p,\sigma}$ will contain the following 
the following set of constant symbols:\\
(1) $0,1,\ldots, p(n)$,\\
(2) $s$, for each $s\in S$ (Note two constants $s_0$ (for initial state),
and $f$ (for final state) will be present in every extensional database),\\
(3) $B$ (blank symbol) and $x$ for each $x \in \Sigma$, and \\
(4) $r,l,\lambda$.\\
We let $edb_{M,\sigma, p}$ consist of the
following set of facts that describe the machine $M$, the segment of
integers $0,\ldots,p(n)$ and the initial configuration $\sigma$ of the tape.
\begin{enumerate}
\item For each $s\in Q$, the clause \ $state(s) \leftarrow$ \
belongs to $\mathit{ext}_{M,p,\sigma}.$
\item For each $x \in \Gamma$, the clause \ $symb(x) \leftarrow$ \
belongs to $\mathit{ext}_{M,p,\sigma}.$
\item For every pair $(s,x) \in Q \times \Gamma$ and every
triple $(s1,x1,d) \in \delta(s,x)$, the clause \ 
$delta(s,x,s1,x1,d) \leftarrow$ \ 
belongs to $\mathit{ext}_{M,p,\sigma}.$
\item For $0 \leq i < p(n)$, the clause \ 
$succ(i,i+1) \leftarrow$ \
belongs to $\mathit{ext}_{M,p,\sigma}.$
\item For $0 \leq i \leq p(n)$, the clause \  
$time(i) \leftarrow$ \ belongs to $\mathit{ext}_{M,p,\sigma}.$
\item For $0 \leq i \leq p(n) -1$, the clause \ $cell(i) \leftarrow$ \ 
belongs to $\mathit{ext}_{M,p,\sigma}.$
\item For $0 \leq m \leq |\sigma | -1$, the clause \ 
$\mathit{data}(m,\sigma (m)) \leftarrow$ \
belongs to $\mathit{ext}_{M,p,\sigma}.$
\item For $|\sigma | \leq m \leq p(n) -1$, the clause \  
$\mathit{data}(m,B) \leftarrow$ \ 
belongs to $\mathit{ext}_{M,p,\sigma}.$
\item The clauses \ $dir(l) \leftarrow$, $dir(r) \leftarrow$  
and $dir (\lambda ) \leftarrow$ \ 
belong to $\mathit{ext}_{M,p,\sigma}.$
\item The clause \ $i\_position (0) \leftarrow$ \ 
belongs to $\mathit{ext}_{M,p,\sigma}.$
\item For all $a,b \in S \cup \Gamma \cup
\{0, \ldots, p(n)\}$ with $a \neq b$, the clause \ 
$neq(a,b) \leftarrow$ \ 
belongs to $\mathit{ext}_{M,p,\sigma}.$
\item For all $a \in S \cup \Gamma \cup
\{0, \ldots, p(n)\}$, the clause \ 
$eq(a,a) \leftarrow$ \ 
belongs to $\mathit{ext}_{M,p,\sigma}.$
\end{enumerate}
The remaining predicates of $P_{\mathit{Trg}}$ are the following:\\
$tape(P,Q,T) $ for ``the tape stores symbol $Q$ at cell $P$
at time $T$'',\\
$position(P,T)$ for ``the read-write head reads the content cell $P$
at time $T$'',\\
$\mathit{state}(S,T)$ for ``the read-write head is in state $S$ at time $T$''
(notice that we have both a unary predicate $\mathit{state}/1$
with the content consisting of states, and $\mathit{state}/2$ to
describe the evolution of the machine),\\
$instr(S,Q,S1,Q1,D,T)$ for ``instruction $(S1,Q1,D)$ belonging to
$\delta (S,Q)$ has been selected for execution at time $T$'',\\
$otherInstr(S,Q,S1,Q1,D,T)$ for ``instruction other than $(S1,Q1,D)$
belonging to $\delta (S,Q)$ has been selected for execution at time $T$'',\\
$\mathit{instr\_def}(T) $ for ``there is an instruction to
be executed at time $T$'',\\
$completion$ for ``computation successfully completed'', and \\
$A$, a propositional letter\footnote{The
propositional letter $A$ will be used whenever we write
clauses acting as constraints. That is, the symbol $A$ will occur in
the following syntactical configuration. $A$ will be the head of
some clause, and the negation of $A$ will also occur in the body
of that same clause. In such situation a stable model {\em cannot}
satisfy the remaining atoms in the body of that clause.}.\\
In the program $P_{\mathit{Trg}}$, there should be no
constants. For notational convenience, we will not be absolutely strict in this respect. That is, to simplify our presentation, 
we will use the constants 0, $f$, and $s_0$ in 
$P_{\mathit{Trg}}$.
These can easily be eliminated by introducing appropriate unary predicates.
Finally to
simplify the clauses, we will
follow here the notation used in the {\em smodels} syntax.
That is, we will use $p(X_1;\ldots;X_k)$ as an abbreviation
for $p(X_1),\ldots, p(X_k)$. 
This given, we are now ready to write the program $P_{\mathit{Trg}}$.

\begin{enumerate}
\item[Group 1.] Our first four clauses are used to describe the position
of the read-write head at any given time $t$.
\begin{enumerate}
\item[(1.1)] (Initial position of the read-write head)\\
$position(P,T) \leftarrow $
$\mathit{time}(T), \mathit{cell}(P)$,
$\mathit{eq}(T,0), i\_position (P)$
\item[(1.2)] 
$position (P1,T1) \leftarrow \mathit{time}(T;T1),$
$\mathit{cell}(P;P1)$,
$ \mathit{state}(S;S1)$,
$\mathit{dir}(D)$,\\
$\mathit{symb}(Q;Q1)$,
$T1 = T+1, P1 +1 = P,$
$position(P,T),$
$\mathit{state}(S,T)$,\\
$\mathit{tape}(P,Q,T),$
$\mathit{instr}(S,Q, S1,Q1,D,T)$,
$D = l, P \neq 0$
\item[(1.3)] $position (P1,T1) \leftarrow$
$\mathit{time}(T;T1)$,
$\mathit{cell}(P;P1)$,
$\mathit{state}(S;S1)$,
$\mathit{dir}(D)$,\\
$\mathit{symb}(Q;Q1)$,
$ T1= T+1,$
$P1 = P +1$,
$position(P,T),$
$\mathit{state}(S,T)$, $\mathit{tape}(P,Q,T),$
$instr(S,Q, S1,Q1,D,T)$, $D = r$, $P\neq p(n)-1$
\item[(1.4)] $position (P1,T1) \leftarrow$
$\mathit{time}(T;T1)$, $\mathit{cell}(P;P1)$, $\mathit{state}(S;S1)$,\\
$\mathit{symb}(Q;Q1)$,
$\mathit{dir}(D)$,
$ T1 = T+1, P = P1, position(P,T)$,\\ 
$\mathit{state}(S,T)$, $\mathit{tape}(P,Q,T),$
$instr(S,Q, S1,Q1,D,T)$, $D = \lambda$
\end{enumerate}
\item[Group 2.] Our next three clauses describe how the contents of the
tape change as instructions get executed.
\begin{enumerate}
\item[(2.1)] $tape(P,Q,T) \leftarrow$
$\mathit{time}(T), \mathit{cell}(P)$, $\mathit{symb}(Q)$,
$T = 0, data(P,Q)$
\item[(2.2)] $tape(P,Q1,T1) \leftarrow$
$\mathit{time}(T;T1)$,
$\mathit{cell}(P)$,
$\mathit{state}(S;S1)$,
$\mathit{symb}(Q;Q1)$,
$\mathit{dir}(D)$,
$T1 = T+1, position(P,T),$
$\mathit{state}(S,T)$, $\mathit{tape}(P,Q,T),$ \\
$instr(S,Q, S1,Q1,D,T)$
\item[(2.3)] $tape(P,Q,T1) \leftarrow$
$\mathit{time}(T;T1)$,
$\mathit{cell}(P;P1)$,
$\mathit{symb}(Q)$,
$ T1 = T+1,$ \\
$ tape(P,Q,T),$
$\mathit{position}(P1,T), P \neq P1$
\end{enumerate}
\item[Group 3.] Our next two clauses describe how the state of the
read-write head evolves in time.
\begin{enumerate}
\item[(3.1)] $state(S,T) \leftarrow$
$\mathit{time}(T)$,
$\mathit{state}(S)$,
$ T = 0, S = s_0$
\item[(3.2)] $state(S1,T1) \leftarrow$
$\mathit{time}(T;T1)$,
$\mathit{cell}(P)$,
$\mathit{symb}(Q;Q1)$,
$\mathit{state}(S;S1)$,\\
$\mathit{dir}(D)$,
$ T1 = T+1, position(P,T),$
$\mathit{state}(S,T)$, $\mathit{tape}(P,Q,T),$\\
$instr(S,Q, S1,Q1,D,T)$
\end{enumerate}
\item[Group 4.] Our next two clauses describe how we select a unique
instruction to be executed at time $T$.
\begin{enumerate}
\item[(4.1)] Selecting instruction at step 0.\\
$instr(S,Q,S1,Q1,D,T) \leftarrow state(S;S1), symb(Q;Q1), dir(D),$\\
$time(T), T= 0$,
$S = s_0, i\_position(P), tape(P,Q,T)$,\\
$delta(S,Q,S1,Q1,D),\ 
\neg otherInstr(S,Q,S1,Q1,D, T)$
\item[(4.2)] Selecting instruction at other steps.\\
$instr(S,Q,S1,Q1,D,T) \leftarrow state(S;S1), symb(Q;Q1), $\\
$dir(D), time(T), T \neq 0, position (P,T), state(S,T),tape(P,Q,T),$\\
$delta(S,Q,S1,Q1,D),\neg otherInstr(S,Q,S1,Q1,D,T)$
\end{enumerate}
\item[Group 5.] Our next set of clauses defines the $\mathit{otherInstr}$
predicate.  Here clauses (5.6) and (5.7) are designed to ensure that
exactly one instruction is selected for
execution at any given time $T$.
\begin{enumerate}
\item[(5.1)] $\mathit{otherInstr}(S,Q,S1,Q1,D1,T) \leftarrow
state(S;S';S1;S2),$\\
$symb(Q;Q';Q1;Q2),$ $time(T),$
$dir(D1;D2),$\\ 
$instr(S',Q',S2,Q2,D2,T),$ $S2 \neq S1$
\item[(5.2)] $\mathit{otherInstr}(S,Q,S1,Q1,D1,T) \leftarrow
state(S;S';S1;S2),$ \\ $symb(Q;Q';Q1;Q2),$ $time(T),$
$dir(D1;D2),$\\ $instr(S',Q',S2,Q2,D2,T),$ $Q2\neq Q1$
\item[(5.3)] $\mathit{otherInstr}(S,Q,S1,Q1,D1,T) \leftarrow
state(S;S';S1;S2),$\\
$symb(Q;Q';Q1;Q2),$ $time(T),$
$dir(D1;D2),$\\
$instr(S',Q',S2,Q2,D2,T),$ $D2\neq D1$
\item[(5.4)] $\mathit{otherInstr}(S,Q,S1,Q1,D1,T) \leftarrow
state(S;S';S1;S2),$\\
$symb(Q;Q';Q1;Q2),$ $time(T),$
$dir(D1;D2),$\\
$instr(S',Q',S2,Q2,D2,T),$ $S' \neq S$
\item[(5.5)] $\mathit{otherInstr}(S,Q,S1,Q1,D1,T) \leftarrow
state(S;S';S1;S2),$\\
$symb(Q;Q';Q1;Q2),$ $time(T),$
$dir(D1;D2),$\\
$instr(S',Q',S2,Q2,D2,T), Q'\neq Q$
\item[(5.6)] The definition of the $\mathit{instr\_def}$ predicate.\\
$\mathit{instr\_def}(T) \leftarrow state(S;S1), symb(Q;Q1),dir(D),
time(T),$ \\
$instr(S,Q,S1,Q1,D,T)$
\item[(5.7)] The clause to ensure that there is an instruction to be
executed at any given time.\\
$A \leftarrow time(T), \neg \mathit{instr\_def} (T), \neg A$
\end{enumerate}
\item[Group 6.] Constraints for the coherence of the computation process.
\begin{enumerate}
\item[(6.1)] When the task is completed.\\
$\mathit{completion} \leftarrow \mathit{symb}(Q),
instr(f,Q,f,Q,\lambda ,p(n))$. 
\item[(6.2)] The atom {\em completion} belongs to every stable model.\\
$A \leftarrow \neg \ \mathit{completion}, \neg A$
\end{enumerate}
\end{enumerate}

Notice that the program $P_{\mathit{Trg}}$ is {\em domain-restricted}
\cite{syr01}, that is, 
every variable in the body of a clause is bound by a positive occurrence
of an extensional database predicate. This restriction does not limit the
expressive power of such  programs, but greatly reduces the work of the
grounding engine \cite{ns97}.

\section{Main Results}\label{results}
Our first proposition immediately follows from our construction.
\begin{proposition}
There is a polynomial $q$ so that for every machine $M$, polynomial $p$,
and an input $\sigma$, the size of the extensional database
$edb_{M,p,\sigma}$ is less than or equal to $q(|M|,|\sigma|, p(|\sigma|))$.
\end{proposition}
We shall now prove that for any
nondeterministic Turing Machine $M$, runtime polynomial $p(x)$, and
input $\sigma$ of length $n$, the stable models of
$edb_{M,p,\sigma} \cup P_{\mathit{Trg}}$ encode the sequences of tapes of
length $p(n)$ which occur in the steps of an accepting computation of
$M$ starting on $\sigma$ and that any such sequence of steps can be
used to produce a stable model of $edb_{M,p,\sigma} \cup P_{\mathit{Trg}}$.

\begin{theorem}\label{propo-main}
The mapping of Turing machines to DATALOG$^\neg$ programs defined by
$\langle M,\sigma ,p\rangle \mapsto edb_{M,p,\sigma} \cup P_{\mathit{Trg}}$
has the property that
there is a 1-1 polynomial time correspondence between the set of
stable models of $edb_{M,p,\sigma} \cup P_{\mathit{Trg}}$ and the set of
computations of $M$ of the length $p(n)$, starting on the tape corresponding 
to the input $\sigma$, and ending in the state $f$.
\end{theorem}
Proof: We first need to see what is a valid run of a machine
$M$ that ends in the state $f$. To this end let us define an {\em
instruction} of the machine $M$ as a quintuple $\la q, a, q_1, a_1,
d\ra$ such that $(q_1, a_1, d) \in \delta (q,a)$. A {\em state of tape} is a
sequence $S$ of symbols of the length $p(n)$ from alphabet $\Sigma \cup \{B\}$. A {\em configuration} is
a triple $\la i,S,k\ra$ where $i$ is an instruction $\la q, a, q_1, a_1,
d\ra$, $S$ is a state of tape, and $k$ is an integer $\leq p(n)$ and
such that $S(k) = a \in \Sigma \cup\{B\}$. Informally, $k$ is the index of the
cell on which the read-write head is pointing at the time the
configuration is observed and $a =
S(k)$ is the content of that cell. The {\em coherence} condition
$S(k) = a$ says that the instruction $i$ is applicable at this moment.
A {\em one step-transition} is a pair of configurations
\[
\la\la i,S,k\ra , \la j,T,m\ra\ra
\]
where $ i = \la q, a, q_1, a_1, d\ra$ and $j = \la r, b, r_1, b_1,
e\ra$ satisfy the following:
\begin{enumerate}
\item $r = q_1$ (i.e. in the transition we moved to the next state as
required by $i$),
\item $m =\left \{ \begin{array}{ll}
k-1 & \mbox{if $k\neq 0, d = l$}\\
k+1 & \mbox{if $k \neq p(n) -1, d = r$}\\
k & \mbox{if $d = \lambda$}
\end{array}\right.$\\
(i.e. the read-write head moved as required by the instruction $i$),
\item $T(n) = \left \{ \begin{array}{ll}
a_1 & \mbox{if $n =k$}\\
S(n) & \mbox{otherwise}
\end{array}\right.$\\
(i.e. the state of tape has been altered in just one place, namely $k$
and $a_1$ has been put there), and 
\item $b = T(m)$. (i.e. the instruction $j$ is coherent with the cell
observed by the read-write head).
\end{enumerate}
We write $C\vdash D$ when $C$ and $D$ are configurations and $\la
C,D\ra$ is a one-step transition.
A {\em run} of a machine $M$ is a sequence of configurations
$\la C_0,\ldots,C_{p(n)}\ra$ such that
\begin{enumerate}
\item $C_0 = \la\la s_0,a,t,a_1,d\ra ,S, 0\ra$ (that is the machine $M$
is in the start state $s_0$, $a = S(0)$ is the content of the cell $0$,
and the read-write head points to cell $0$) and 
\item for all $0\leq k < p(n)$, $C_k \vdash C_{k+1}$.
\end{enumerate}
A {\em valid run} of the machine $M$ is a run where $C_{p(n)} =
\la i,S,k\ra$, $i = \la f,a,f,a,\lambda\ra$. Thus a valid run of $M$ is
a run where the last state of the machine is $f$. 

For the rest of this proof, we shall only consider valid runs 
$\la C_0,\ldots,C_{p(n)}\ra$ of $M$ such that 
$C_0 = \la i, S, 0 \ra$ where $S(i) = \sigma(i)$ 
for $i \leq n-1$ and $S(i) = B$ for $i > n$.  That is, we shall only 
consider valid runs of $M$ which start on an  
input $\sigma = (\sigma(0), \ldots, \sigma(n-1))$ of length $n$.  
We will show that each 
such valid run determines a unique stable
model of $edb_{M,p,\sigma} \cup P_{\mathit{Trg}}$ and conversely
every stable model of the program
$edb_{M,p,\sigma} \cup P_{\mathit{Trg}}$ determines such a  valid run of $M$.
First, given a valid run ${\cal C} = \langle C_0,\ldots ,C_{p(n)}\rangle$ of
the machine $M$, where for $m$, $0 \leq m \leq p(n)$,
\[
C_m = \langle i_m, S_m, k_m\rangle
\]
we define the set of atoms $N_{\cal C}$ which consists of the union of
sets of atoms $N_1\cup \ldots\cup N_7$ where:
\begin{description}
\item $N_1 = edb_{M,p,\sigma}$
\item $N_2 = \{ \mathit{position}(m,k_m) : 0\leq m \leq p(n)\}$
\item $N_3 = \{ \mathit{tape}(r,S_m(r), m) : 0 \leq m \leq p(n), \
0 \leq r \leq p(n)-1 \}$
\item $N_4 = \{\mathit{instr}(q,a,q',a',d, m) : i_m = \langle
q,a,q',a',d\rangle, 0 \leq m \leq p(n)\}$
\item $N_5 = \{ \mathit{otherInstr}(q'',a'',q''',a''',d''',m) : 
\langle q''',a''',d'''
\rangle \in \delta (q'',a''),$\\
$\left. \ \ \ \ \ \ \ i_m = \langle q,a,q',a',d\rangle , 
\langle q'', a'', q''', a''', d''' \rangle \neq \langle q, a, q',a',d\rangle , 0\leq m \leq
p(n)\} \right.$
\item $N_6 = \{ \mathit{instr\_def}(m) : 0 \leq m \leq p(n) \}$
\item $N_7 = \{\mathit{completion}\}$
\end{description}
We show that $N_{\cal C}$ is a stable model of $edb_{M,p,\sigma}
\cup P_{\mathit{Trg}}$.
Indeed, after we ground $edb_{M,p,\sigma} \cup P_{\mathit{Trg}}$ and reduce it
with respect to $N_{\cal C}$,  it is straightforward to prove by simultaneous induction on $m$
that the least model of the
reduct contains
\begin{enumerate}
\item[(a)] for each $m$, $0\leq m \leq p(n)$ exactly one
atom of the form\\
$\mathit{instr}(q,a,q',a',d, m)$ and that this atom
belongs to $N_{\cal C}$,
\item[(b)] for each $m$, $0 \leq m\leq p(n)$ all atoms
$\mathit{otherInstr}(q'', a'', q''',a''', d''', m)$
with $\langle q''',a''',d''' \rangle \in \delta (q'',a'')$, where
$i_m = \langle q,a,q',a',d\rangle$,
and 
\[ \langle q'', a'', q''', a''' ,d''' \rangle \neq \langle q, 
a, q, a',d\rangle,
\]
and that these atoms belong to $N_{\cal C}$,
\item[(c)]
for each $m$ and $r$, $0 \leq m \leq p(n)$, $0 \leq r \leq p(n) -1$
exactly one atom of the form $\mathit{tape}(r,x,m)$
that these atoms belong to $N_{\cal C}$
\item[(d)] for each $m$, $0\leq m \leq p(n)$ exactly one atom of the form
$\mathit{position}(m,k)$ and that $k = k_m$,
\item[(e)] for each $m$, $0\leq m \leq p(n)$ exactly one atom of the form
$\mathit{instr\_def}(m)$, and
\item[(f)] the atom {\em completion}.
\end{enumerate}

Thus $N_{\cal C}$ is a stable model of 
$edb_{M,p,\sigma} \cup P_{\mathit{Trg}}$.
Moreover, it is clear that the mapping ${\cal C} \mapsto N_{\cal C}$ is
an injection since two different valid runs $\cal C$ and ${\cal C}'$
differ in some least place $m$ and, hence, the atoms of group $N_4$ 
involving the relational symbol {\em instr} at time $m$ must differ in 
$N_{\cal C}$ and $N_{{\cal C}'}$.

Conversely, suppose that $N$ is a stable model of 
$edb_{M,p,\sigma} \cup P_{\mathit{Trg}}$. First observe that $A \notin N$.  
That is, all the clauses that have $A$ in the head also have $\neg A$ 
in the body.  Thus if $A \in N$, then 
there are no clauses with head $A$ in $GL(edb_{M,p,\sigma} 
\cup P_{\mathit{Trg}},N)$ so that $N$ could not be a stable model.  
Since, $A$ is not in $N$, then it is easy to  to see that clause (6.2) 
forces $N$ to contain the atom {\em completion}. Since the only way 
to derive the atom {\em completion} is via clause (6.1), 
it follows that $N$ must contain the atom 
$instr(f,q,f,q,\lambda,p(n))$ for some symbol $q \in \Gamma$.

Similarly, by clause (5.7), for every $0 \leq t \leq p(n)$, it must 
be the case that $\mathit{instr}\_{def}(t)$ must be in $N$. Since the only way 
to derive $\mathit{\mathit{instr}\_def}(t)$ is via clause (5.6), it 
follows that for each $0 \leq t \leq p(n)$, there must exist 
$s, q, s1,q1$ and $d$ such that $instr(s,q,s1,q1,d,t) \in N$. 
There cannot be a time $t$ with $0 \leq t \leq p(n)$ such that there 
two different 6-tuples $(s,q,s1,q1,d,t)$ and 
$(s',q',s1',q1',d',t)$ such that both atoms $instr(s,q,s1,q1,d,t)$ and 
$instr(s',q',s1',q1',d',t)$ are in $N$ because then it follows from clauses 
(5.1)-(5.5), that $otherInstr(s2,q2,s3,q3,d3,t)$ holds for all 
5-tuples in $S \times \Gamma \times S \times \Gamma \times D$ such 
that $(s3,q3,d3) \in \delta(s2,q2)$.  
But then the only clauses that have $instr(s,q,s1,q1,d,t)$ in the 
head are the clauses in either (4.1) or (4.2) and all such clauses  would 
all be eliminated in the construction of 
$GL(edb_{M,p,\sigma} \cup P_{\mathit{Trg}},N)$ so that there  
could be no  $instr(s,q,s1,q1,d,t)$ in $N$. Thus it follows that for each 
$0 \leq t \leq p(n)$, there is a unique $(q_t,a_t,q_t',a_t',d_t)$ such that 
$instr(q_t,a_t,q_t',a_t',d_t,t)$ is in $N$.  

It is then easy to check that our clauses in groups (1) through (4) 
ensure that the instructions 
$\{(q_t,a_t,q_t',a_t',d_t): 0 \leq t \leq p(n)\}$ determine a valid run of 
the Turing machine $M$ started on input $\sigma$. In particular, for 
each $0 \leq t \leq p(n)$, there is a unique position $p_t$ such 
that the atom $\mathit{position}(p_t,t)$ is in $N$ and 
$q_t$ is the only state such 
that $state(q_t,t)$ is in $N$ and $a_t$ is the only symbol such that 
$tape(p_t,a_t,t)$ is in $N$.  Moreover it is easy to check that for each 
time $0 \leq t \leq p(n)$  and each cell $0 \leq c \leq p(n)-1$, there is 
a unique symbol $a_{t,c}$ such that $tape(c,a_{t,c},t)$ is in $N$.
It follows if we define the sequence
\[
{\cal C}_N = \langle C_0,\ldots ,C_{p(n)}\rangle
\]
so that for each $0 \leq t \leq p(n)$, $C_t = \langle i_t, S_t, k_t\rangle$ 
where 
\begin{enumerate}
\item $i_t = \langle q_t,a_t,q'_t,a'_t,d_t\rangle$ and $\mathit{instr}
(q_t,a_t,q'_t,a'_t,d_t,t) \in N$, 
\item $S_t = \{<r,a> : \mathit{tape}(r,a,t) \in N\}$ and 
\item $k_t$ is the only $k$ such that $\mathit{position}(k,t) \in N$,
\end{enumerate}
then ${\cal C}_N$ is a valid run of $M$. Finally,
it is easy to show by induction that 
$N_{{\cal C}_N} = N$. This, together with
the fact that ${\cal C} \mapsto N_{\cal C}$ is one-to-one completes our
argument. \hfil $\Box$
\begin{corollary}\label{FF1}
A search problem $\cal S$ can be solved by means of a uniform logic
program in SLP if and only if $\cal S$ is an $\mathit{NP}$-search problem.
\end{corollary}

A {\em supported model} $M$ of a propositional logic program $P$ is a subset 
of the Herbrand base of $P$ that is a fixed point of the one step 
provablility operator, $T_P$ associated with $P$. That is, $M$ is 
a supported model of $P$ if and only if $M = T_P(M)$. Thus a supported model 
$M$ consists precisely of heads of clauses with bodies satisfied by $M$.

Given a logic program $P$ we say that $M$ is supported model of 
$P$ if and only if $M$ is a supported model of $P_g$. 

\begin{lemma}\label{FF2} 
For all $M$, $p$, and $\sigma$, $M$ is a supported model of 
$edb_{M,p,\sigma} \cup P_{\mathit{Trg}}$ if and only if $M$ is stable 
model of $edb_{M,p,\sigma} \cup P_{\mathit{Trg}}$. 
\end{lemma}
Proof: It is well known that for every program $P$,
every stable model of $P$ is a supported model of $P$. 

For the other direction, suppose 
that $R = edb_{M,p,\sigma} \cup P_{\mathit{Trg}}$ and suppose 
that $M$ is a supported model of $R$ so that $T_R(M) = M$.  First we 
observe that $A$ cannot be in $M$.
That is, all the clauses that have $A$ in the head also have $\neg A$ 
in the body.  Thus if $A \in M$, then 
there are no clauses with head $A$ whose body is satisfied 
by $M$ and hence  $A$ would not be in $T_R(M)$ which would violate 
that our assumption that $T_R(M) =M$.   
Since $A$ is not in $M$, then it is easy to see that clause (6.2) 
forces $M$ to contain the atom {\em completion} since otherwise 
$A$ would be in $T_R(M)$. 
Similarly, by clause (5.7), for every $0 \leq t \leq p(n)$, it must 
be the case that $\mathit{instr}\_{def}(t)$ must be in $M$. Since the only way 
to derive $\mathit{\mathit{instr}\_def}(t)$ is via clause (5.6), it 
follows that for each $0 \leq t \leq p(n)$, $M$ must satisfy the body 
of clause (5.6) where $T =t$. Thus 
there must exist 
$s, q, s1,q1$ and $d$ such that $instr(s,q,s1,q1,d,t) \in M$. 
We claim that there 
cannot be a time $t$ with $0 \leq t \leq p(n)$ such that there 
two different 6-tuples $(s,q,s1,q1,d,t)$ and 
$(s',q',s1',q1',d',t)$ such that both atoms $instr(s,q,s1,q1,d,t)$ and 
$instr(s',q',s1',q1',d',t)$ are in $M$. Otherwise the clauses 
(5.1)-(5.5) would show that that $otherInstr(s2,q2,s3,q3,d3,t) \in T_R(M) = M$ 
for all 5-tuples in $S \times \Gamma \times S \times \Gamma \times D$ such 
that $(s3,q3,d3) \in \delta(s2,q2)$.  
But then the only clauses that have $instr(s,q,s1,q1,d,t)$ in the 
head are the clauses in either (4.1) or (4.2) and the body of all 
such clauses would not be satisfied by $M$. Hence  there  
could be no  $instr(s,q,s1,q1,d,t)$ in $T_R(P)$ which would 
violate our assumption that $T_R(M) = M$. 
Thus it follows that for each 
$0 \leq t \leq p(n)$, there is a unique $(q_t,a_t,q_t',a_t',d_t)$ such that 
$instr(q_t,a_t,q_t',a_t',d_t,t)$ is in $M$.  

We can then proceed exactly as in our proof of Theorem \ref{propo-main} to 
prove by induction that the fact $T_R(M) = M$ 
implies that $M$ must be of the form 
$N_{\cal C}$ where ${\cal C} = \langle C_0,\ldots ,C_{p(n)}\rangle$ is a 
valid run of the machine $M$ started on input $\sigma$. Thus 
$M$ must be a stable model of $M$. $\hfill\Box$

Lemma \ref{FF2} implies that analogue of Corollary \ref{FF1} holds for Supported Logic Programming, SuLP.
\begin{corollary}
A search $S$ problem can be solved by means of a uniform logic
program in SuLP if and only if $S$ is an $\mathit{NP}$-search problem.
\end{corollary}

We can also prove similar results for default logic programs
without function symbols with respect to nondeterministic Turing machines
with an oracle for 3-$\mathit{SAT}$.

\begin{theorem}\label{thm-defo}
For each $n\in N$ there is a default theory
$\langle W_n,D_n\rangle$ such that for
every 3-SAT oracle Turing machine $M$, every polynomial $p \in N[x]$,
and every finite input $\sigma$ where $ |\sigma | = n$, there is a
polynomial-time one-to-one correspondence between the accepting
computations of length $p(n)$ of $M$ on input $\sigma$ and the Reiter
extensions of the default theory $\langle \mathit{edb}_{M,p,\sigma}
\cup W_n, D_n \rangle$.
\end{theorem}
The proof of this result is more involved and requires additional
technical means. It will be a subject of a separate publication. 
It follows from Theorem \ref{thm-defo} 
that a search problem $S$ can be solved by means of a
uniform default logic program if and only if $S$ is in $\Sigma_2^P$. A
version of this result for decision problems has been proved in \cite{ceg97}.
\section*{Acknowledgements}
The first author's research has been 
partially supported by the NSF grant IIS-0097278\footnote{Any opinions, 
findings, and
conclusions or recommendations expressed in this material are those of
the authors and do not necessarily reflect the views of the National
Science Foundation.}.
The second author's research has been 
partially supported by the ARO contract DAAD19-01-1-0724.

\newcommand{\etalchar}[1]{$^{#1}$}
\bibliographystyle{acmtrans}

\end{document}